\documentclass{article}
\usepackage{spconf,amsmath,graphicx}
\usepackage{mathrsfs, amssymb}
\usepackage{array}
\usepackage{cite}

\DeclareMathOperator*{\argmax}{arg\,max}
\DeclareMathOperator*{\softmax}{Softmax}

\title{A SEQUENTIAL GUIDING NETWORK WITH ATTENTION FOR IMAGE CAPTIONING}
\name{Daouda Sow$^{1}$ \qquad Zengchang Qin$^{1,2}$ \qquad Mouhamed Niasse$^{3}$ \qquad Tao Wan$^{1}$ }
\address{$^{1}$ Intelligent Computing and Machine Learning Lab, School of ASEE, Beihang University, China\\  $^{2}$ Keep Labs, Keep Inc., Beijing, China 
$^{3}$ School of EEE, North China Electric Power University, China\\
{\small \texttt{\{davissow, zcqin, taowan\}@buaa.edu.cn}}
}

\begin{document}
\ninept
%
\maketitle
\begin{abstract}
The recent advances of deep learning in both computer vision (CV) and natural language processing (NLP) provide us a new way of understanding semantics, by which we can deal with more challenging tasks such as automatic description generation from natural images. In this challenge, the encoder-decoder framework has achieved promising performance when a convolutional neural network (CNN) is used as image encoder and a recurrent neural network (RNN) as decoder. In this paper, we introduce a sequential guiding network that guides the decoder during word generation. The new model is an extension of the encoder-decoder framework with attention that has an additional guiding long short-term memory (LSTM) and can be trained in an end-to-end manner by using image/descriptions pairs. 
We validate our approach by conducting extensive experiments on a benchmark dataset, i.e., MS COCO Captions. The proposed model achieves significant improvement comparing to the other state-of-the-art deep learning models. 
\end{abstract}
%
%
\section{Introduction}
\label{sec:intro}

Automatically describing the content of images is of one of the hardest tasks in scene understanding \textemdash a long standing problem of the field of artificial intelligence (AI). This task is challenging as it requires semantic understanding both of image and natural language, building a good learning models to shrink semantic gaps in different modalities. This unsolved problem has attracted a lot of attentions in the AI community \cite{Vinyals2017ShowAT,Karpathy,Xu2015ShowAA, Yao2017BoostingIC, Jia2015GuidingLT, Jiang2018LearningTG, Chen2015MindsEA, Kiros2014UnifyingVE, capFang, Mao2014DeepCW, Donahue2015LongTermRC, Liu2017MATAM, Chen2017SCACNNSA, Bernardi2017AutomaticDG}. 
An image captioning model must detect all objects/scenes present in an image and sometimes even objects/scenes that are not present in the image but related to its description. Moreover, it should be able to capture the relations between the objects/scenes and express them in a well-formed, human-understandable sentence. This challenge has the significance not only in academic research, but also in various applications such as information retrieval and visual question-answering.  

Some recent research of image captioning take inspiration from neural machine translation systems (NMT) \cite{Cho2014LearningPR, MT2, Sutskever2014SequenceTS, Luong2015EffectiveAT} that successfully use sequence-to-sequence learning for translation. NMT models solve the task of translation by a two-fold pipeline. (1) A RNN is used to encode the source sentence in a fixed-length vector and then (2) a decoding RNN is  conditioned on that fixed-length vector to generate a sentence in the target language, one word at each time step. This general encoder-decoder framework is well suited to the image description problem as the task is (sort of) equivalent to translating an image into its corresponding description. 
However, image captioning appears to be much harder than certain machine translation tasks. For instance, when translating from English to French, the source and target languages share similar sentence structures (similar part-of-speech order). This similarity in structure is very useful for the translating system as the alignment will be much easier. Instead, for image captioning, the structure of the visual data is way different from the structure of the captions describing them. Moreover, the simple CNN+RNN pipeline squash the whole input image into a fixed-length embedding vector. This constitutes a major limitation of the basic encoder-decoder architecture. 

To overcome these limitations both for machine translation and image captioning, some new models were proposed by using the \emph{attention} mechanism \cite{MT2, Xu2015ShowAA, Luong2015EffectiveAT}. For image captioning, attention mechanisms help the model focus on salient regions of the image while generating descriptions by dynamically updating the image representations at each time step. With this the input image is now represented as a sequence of context vector where the length of the sequence depends on the number of words in the sentence to be generated. Promising results has been published since attention was introduced in \cite{MT2} then later refined in \cite{Luong2015EffectiveAT}. Another group of models \cite{Yao2017BoostingIC, Jia2015GuidingLT, Jiang2018LearningTG} tried to overcome the limitation of the basic encode-decoder framework by still representing the input image as a fixed-length vector but injecting external guiding information at each time step. The external guiding information can be any attribute features connecting the image to its description. For instance, the attribute features could be semantic information obtained from a multimodal space of images and their descriptions learned using Canonical Correlation Analysis \cite{Jia2015GuidingLT} or the prediction of frequent word occurrences in captions \cite{Yao2017BoostingIC} or even learned by an additional guiding network \cite{Jiang2018LearningTG}. The guiding information is however static in all of these models and couldn't be adjusted during the process of generation. 
In this work, we investigate how can we take advantages of these encoder-decoder models by constructing a joint neural model with attention that has an extra guiding network.
Our approach is more closely related to the work of \cite{Jiang2018LearningTG} but instead of learning one magic guiding vector, we propose to learn a sequential guiding network that can adapt its guiding vector during words generation. More specifically, the guiding network is a long short-term memory (LSTM) which outputs a guiding vector at each time step based on previous guiding vectors, current attention vector and attribute features. 
We use the Luong style of attention \cite{Luong2015EffectiveAT} which is a refined version of attention mechanism and that to the best of our knowledge, there has not been any published work reporting the performance of an image captioning model that is built following \emph {only} the encoder-decoder pipeline with Luong style of attention.
Furthermore, to demonstrate the usefulness of the guiding LSTM, we also compare the performance of our model with and without the the guiding LSTM in experiments.  

\section{Related work}
\label{sec:rw}

In the past few years, the advances in training deep neural network 
models both for CV \cite{NIPS2012_4824} and NLP \cite{Cho2014LearningPR} give new perspectives for automatic
image description generation. The neural encoder-decoder framework of 
machine translation has recently been used for generating image captions 
because of the high level similarity of the two fields. Both fields aim to translate a source ‘language’ to a target one. The model in \cite{Kiros2014UnifyingVE} was the first to follow the encoder-decoder pipeline for image captioning. 
Authors of \cite{Kiros2014UnifyingVE} use a CNN to compute image features and a LSTM model to encode the corresponding descriptions. 
The image features are projected into latent states of the LSTM encoder to construct a multimodal distributed representation learned by optimizing a simple pairwise ranking loss. Image descriptions were generated from the multimodal space using a novel multiplicative neural language model named Structure-Content Neural Language Model (SC-NLM). Their approach gives superior results than all previous models but was later outperformed by the model described in \cite{Karpathy}, which is a simpler encoder-decoder architecture, again directly inspired by Neural Machine Translation (NMT).  \cite{Karpathy} pass images through a deep CNN, take the activation of the last fully-connected layer as image features and then initialize the hidden states of a RNN cell with the CNN image features. During training at each time step they input the current word, compute a distribution over all the words of the vocabulary based on the hidden 
states and maximize the likelihood of the true next word using a negative log likelihood loss. The work in \cite{Vinyals2017ShowAT} employs a more powerful RNN cell, and they incorporated the image features as first input word instead of using it as initial hidden state. Other similar approaches include \cite{Mao2014DeepCW, Jia2015GuidingLT, Donahue2015LongTermRC}. \cite{Jia2015GuidingLT} were proposed as an extension of the LSTM model by exploring different kind of semantic information that can be used as extra guiding input to the LSTM during decoding steps. \cite{Yao2017BoostingIC} followed this direction by injecting more powerful high-level image attributes into the decoder. In their work, they investigate different architecture for injecting word occurrence prediction attributes \cite{capFang} into the CNN-RNN framework. 
 
Inspired by the use of attention in sequence-to-sequence learning for machine translation \cite{MT2,Luong2015EffectiveAT}, visual attention has been proved to be a very effective way of improving image captioning. 
Some early research follows this direction, e.g., the model proposed in \cite{Xu2015ShowAA} can focus on important parts of images while generating descriptions. The captioning model in \cite{Xu2015ShowAA} is very similar in spirit to that in \cite{Chen2015MindsEA}, in which visual representation is constructed for sentence parts while the description is being emitted. In \cite{Chen2017SCACNNSA}, authors proposed a spatial and channel-wise attention mechanism over 3D CNN feature maps. In addition to spatial attention (standard visual attention), their model can also learn to pay attention over CNN channels, which they argued as extracting semantic attributes. 
In \cite{Jiang2018LearningTG}, visual attention is combined with semantic information for generating image captions. 
Their model can learn an additional guiding vector while learning to focus on image regions. However in contrast to our model, their framework only learns a fix guiding vector that couldn't be adapted during words generation. 

A pure sequence-to-sequence architecture for image captioning is proposed in \cite{Liu2017MATAM}. 
Different from previous approaches, their model represents images as 
a sequence of detected objects and a \emph{'sequential attention layer'} is introduced to help the model focus on important objects. While resulting in a more complex architecture, their approach claims state-of-the-art 
results in all metrics. Instead of training via (penalized) maximum likelihood estimation, some recent works use Policy Gradient (PG) methods to directly optimize the non-differentiable testing metrics, claiming boost in term of performance measure. While \cite{Rennie2017SelfCriticalST} optimize for the standard CIDEr metric, \cite{Liu2017ImprovedIC} proposed to optimize for a new testing metric that is a linear combination of CIDEr \cite{Vedantam2015CIDErCI} and SPICE \cite{Anderson2016SPICESP} they called SPIDEr, which they found better correlated with human judgment. However in this line of work, it is not clear yet whether the improvement in testing metrics could result in captions with better quality. 

\section{Proposed model}
\label{sec:model}

Technically, the ultimate goal of the neural CNN+RNN architecture for image captioning is to build an end-to-end model trainable by standard backpropagation that can generate a description $S^i$ given a certain image $X^i$. Inspired by NMT, such a model can ‘translate’ an image into a describing sentence. A CNN is first used to obtain image features and a RNN decoder is conditioned on those CNN image features to generate the describing sentence. Given a training dataset consisting of $(S^i, X^i)$ pairs, these models aim to directly maximize the log-probability of generating $S^i$ when $X^i$ is the input. Thus, the optimization problem can be formulated by: 
\begin{equation}
\theta^* = \argmax_{\theta} \sum_i {\log p(S^i|X^i; \theta)}
\end{equation}
where $\theta$ represents the set of parameters to be learned, $X^i$ is a single image and $S^i=[w_1^i, w_2^i, ..., w_{N^i}^i]$ is the corresponding caption which is a sequence of $N^i$ words. Because each caption represents a sequence of $N^i$ words, the log probability is calculated using Bayes chain rule:
\begin{equation}
\log p(S^i|X^i; \theta) = \log p(w_1^i|X^i; \theta) + \sum_{i=2}^{N^i} {\log p(w_t^i|X^i, w_{1:t-1}^i; \theta)}
\end{equation}
where $p(w_1^i|X^i; \theta)$ is the likelihood of generating the first word $w_1^i$ given only the image $X^i$ and $p(w_t^i|X^i; w_{1:t-1}^i; \theta)$ represents the probability of emitting word $w_t^i$ at time step $t$ conditioned on the image $X^i$ and the words generated so far $w_{1:t-1}^i$. In our work, we model the distribution $p(w_t^i|X^i, w_{1:t-1}^i; \theta)$ with a LSTM cell wrapped with Luong-style attention mechanism \cite{Luong2015EffectiveAT}. A RNN cell with Luong's attention computes its hidden state ${\boldsymbol h}^t$ at each time step based on the current input ${\boldsymbol x}^t$, the previous hidden state ${\boldsymbol h}^{t-1}$ and the previous attention vector ${\boldsymbol {\tilde h}}^{t-1}$. The current hidden state ${\boldsymbol h}^t$ is combined with the image-side context vector ${\boldsymbol c}^t$ to form the final output of the cell which is the current attention vector ${\boldsymbol {\tilde h}}^t$. Finally, the distribution $p(w_t^i|X^i; w_{1:t-1}^i; \theta)$ is computed by applying a Softmax layer on top of the current attention vector ${\boldsymbol {\tilde h}}^t$: 
\begin{equation}
\begin{gathered}
p(w_t^i|X^i; w_{1:t-1}^i; \theta) = \softmax ({\boldsymbol W}_s{\boldsymbol {\tilde h}}^t) \\
{\boldsymbol {\tilde h}}^t = \tanh ({\boldsymbol W}_c[{\boldsymbol c}^t; {\boldsymbol h}^t]) \\
{\boldsymbol h}^t = \mathscr{R} ({\boldsymbol x}^t) \\
\end{gathered}
\label{eq:1}
\end{equation}
where ${\boldsymbol W}_s$, ${\boldsymbol W}_c$ are projection matrices, ${\boldsymbol c}^t$ is the image-side context vector detailed in section \ref {ssec:cnn} and $\mathscr{R}$ is a recursive function whose details are given in Section \ref {ssec:lstma}.

\begin{figure}[ttb]
\centering
\includegraphics[width=8.5cm, height=4.8cm]{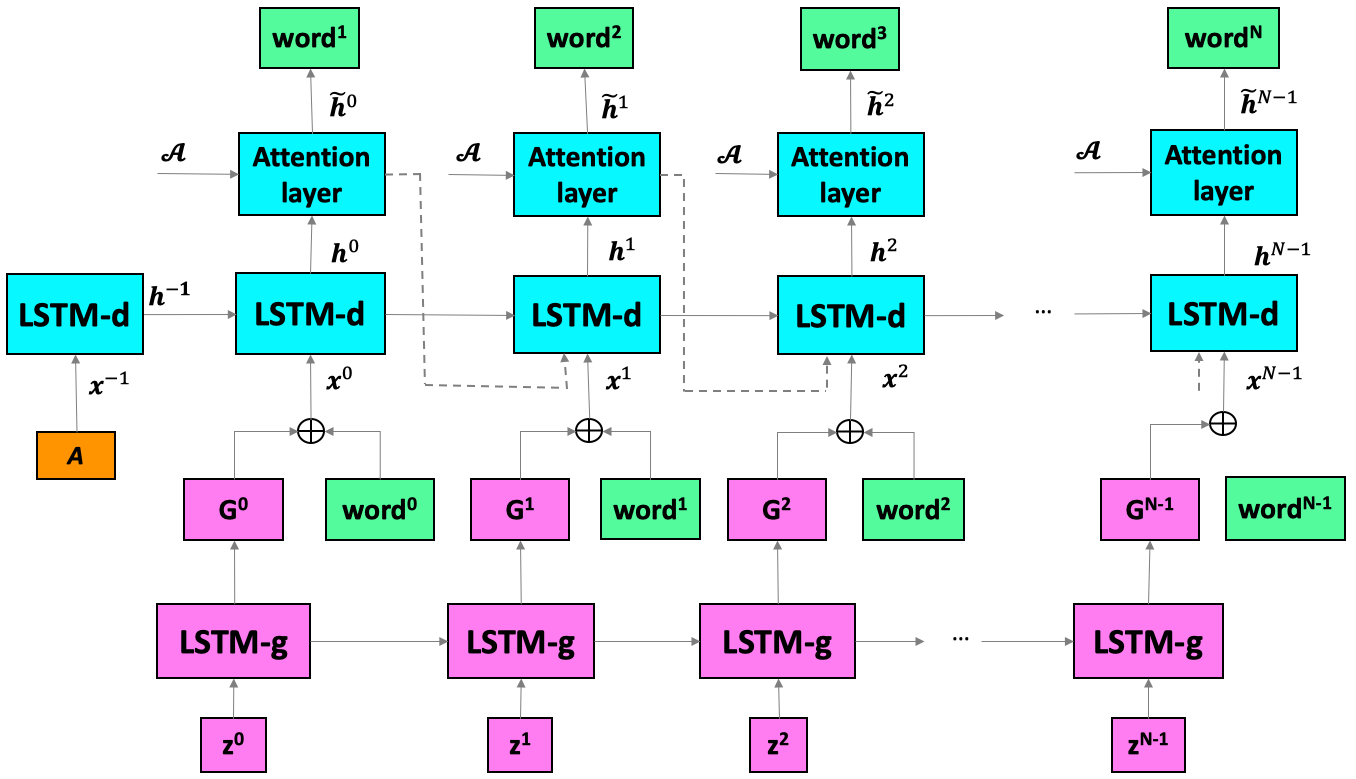}
\caption{Unrolled version of our model (better viewed in color).}
\label{fig:model}
\end{figure}

\subsection{Image Features: Convolutional Neural Network}
\label{ssec:cnn}
Convolutional neural networks (CNNs) are powerful models in Computer Vision with state-of-the-art performance in  image classification and object detection. 
Let $ \mathcal{A}=\{{\boldsymbol a}_1, {\boldsymbol a}_2, ..., {\boldsymbol a}_K\}$ 
denotes the set of annotation vectors (attention states) extracted  from the last 
convolutional layer of the encoding CNN. Here $K$ is the number of neurons in one activation map of the last convolutional layer and 
each ${\boldsymbol a}_k$ is a $D$-dimensional feature vector where $D$ is 
the number of activation maps of the last convolutional layer. With those 
attention states, our model computes a context vector ${\boldsymbol c}^t$ 
which is a weighted sum of attention states and can be seen as a dynamic 
representation of the image at every time step $t$:
\begin{equation}
\begin{gathered}
s_k^t = \Phi({\boldsymbol a}_k, {\boldsymbol h}^t) \\
\alpha_k^t = \frac {\exp(s_k^t )}{\sum_{j=1}^{K} \exp(s_j^t )} \\
{\boldsymbol c}^t = \sum_{k=1}^{K} \alpha_k^t \times {\boldsymbol a}_k
\end{gathered}
\end{equation}
where ${\boldsymbol c}^t \in \mathbb{R}^D$ and 
$\alpha_1^t, \alpha_2^t, \ldots, \alpha_K^t$ are alignment weights 
and the function $\Phi$ is known as alignment function. In our model, 
we use the general form described in \cite{Luong2015EffectiveAT}: 
\begin{equation}
\Phi({\boldsymbol a}_k, {\boldsymbol h}^t) ={ {\boldsymbol h}^t}^\top {\boldsymbol W}_a 
{\boldsymbol a}_k
\end{equation}
where ${\boldsymbol W}_a \in \mathbb{R}^{H\times D}$ is a transformation matrix.

\subsection{Sentence Generator: LSTM + Luong's Attention}
\label{ssec:lstma}
The form of the recursive function $\mathscr{R}$ (Equation \ref{eq:1}) is a critical design choice 
for generating sequences. In this paper, we use a LSTM cell wrapped with the 
attention mechanism described in \cite{Luong2015EffectiveAT} to form $\mathscr{R}$. LSTM \cite{Hochreiter} is a powerful form of recurrent neural network that is widely used now because of its ability to deal with issues like vanishing and exploding gradients. 
The final form of $\mathscr{R}$ is described by the following equations:
\begin{equation}
\begin{gathered}
{\boldsymbol i}^t = \sigma({\boldsymbol W}_{xh}^i{\boldsymbol x}^t + 
{\boldsymbol W}_{hh}^i{\boldsymbol h}^{t-1} + 
{\boldsymbol W}_{{\tilde h}h}^i{\boldsymbol {\tilde h}}^{t-1} + 
{\boldsymbol b}^i) \\
{\boldsymbol f}^t = \sigma({\boldsymbol W}_{xh}^f{\boldsymbol x}^t + 
{\boldsymbol W}_{hh}^f{\boldsymbol h}^{t-1} + 
{\boldsymbol W}_{{\tilde h}h}^f{\boldsymbol {\tilde h}}^{t-1} + 
{\boldsymbol b}^f) \\
{\boldsymbol o}^t = \sigma({\boldsymbol W}_{xh}^o{\boldsymbol x}^t + 
{\boldsymbol W}_{hh}^o{\boldsymbol h}^{t-1} + 
{\boldsymbol W}_{{\tilde h}h}^o{\boldsymbol {\tilde h}}^{t-1} + 
{\boldsymbol b}^o) \\
{\boldsymbol g}^t = \sigma({\boldsymbol W}_{xh}^g{\boldsymbol x}^t + 
{\boldsymbol W}_{hh}^g{\boldsymbol h}^{t-1} + 
{\boldsymbol W}_{{\tilde h}h}^g{\boldsymbol {\tilde h}}^{t-1} + 
{\boldsymbol b}^g) \\
{\boldsymbol m}^t = {\boldsymbol f}^t \odot {\boldsymbol m}^{t-1} + 
{\boldsymbol i}^t \odot {\boldsymbol g}^t \\
{\boldsymbol h}^t = {\boldsymbol o}^t \odot \tanh ({\boldsymbol m}^t)
\end{gathered}
\end{equation}
here ${\boldsymbol x}^t$ is input signal at time step $t$, ${\boldsymbol m}^t$ and ${\boldsymbol h}^t$ are respectively memory cell and hidden state of the LSTM cell and ${\boldsymbol {\tilde h}}^t$ represents attention vector. The variables ${\boldsymbol i}^t, {\boldsymbol f}^t, {\boldsymbol o}^t, {\boldsymbol g}^t$ stand respectively for input gate, forget gate, output gate and candidate memory cell. The various ${\boldsymbol W}_{xy}^z$ and ${\boldsymbol b}^z$ are respectively parameter matrices and bias terms to be optimized. The non-linearity $\sigma$ is element-wise sigmoid activation and $\odot$ is the element-wise dot product.

\subsection{Sequential Guiding Network}
\label{sec:sgn}
While the decoding function can access image features at each time step in the encoder-decoder framework with attention, injecting additional guiding vector to the decoder input signal can lead to higher performance. In our work, we extend the CNN+RNN architecture with attention by inserting an extra guiding network. 
Different from previous approaches that learn a static guiding vector, we explore the use of a sequential guiding network that can adapt its guiding vector at every time step. We model the sequential guiding network with a LSTM cell and name it LSTM-g. In this way, the guiding vector \( \boldsymbol G^{t} \) which is the hidden state of LSTM-g, can be adjusted based on previous guiding vectors and current guiding input signal \( \boldsymbol z^{t} \). We then use the guiding vector  ${\boldsymbol G}^{t}$  to construct the input signal  \( \boldsymbol x^{t} \)  for the decoding cell  \( \mathscr{R} \) . The guiding input signal  \( \boldsymbol z^{t} \)  at every time step is formed by concatenating the previous attention vector and attribute features. Figure \ref {fig:model} shows an unrolled version of our framework. We referred the LSTM in the decoding cell as LSTM-d. 

\subsubsection{High-Level Image Attributes}
\label{sssec:hlim}
In addition to the CNN features, our model also integrates other high-level attributes of the input image in the decoding phase. The probability distribution over most frequent words in captions has been shown to be powerful and very informative for image description \cite{Yao2017BoostingIC,capFang}. We explore the use of this kind of image attributes in our model and denote by  \( \boldsymbol A \in \mathbb{R}^{D_a} \)  the detected attribute representations. However, the high-level image attributes could be any additional attribute features connecting the image to its describing sentence. The attributes vector is used to construct initial state for the decoding LSTM and as additional guiding information for the guiding LSTM. 

\subsubsection{Complete Updating Procedure}
\label{sssec:cup}
Given the input image represented by the set of annotation vectors  \( \mathcal A \)  
and attribute features  \( \boldsymbol A \) and the describing sentence represented 
by the sequence $\left[ \boldsymbol w_{0}^{i}, \boldsymbol w_{1}^{i},  \ldots , 
\boldsymbol w_{N^{i}}^{i} \right]$ , the decoding cell  \( \mathscr R \)  
update its hidden state  \( \boldsymbol  h^{t} \)  at each time step following the procedure: 
\begin{equation}
\begin{gathered}
\boldsymbol x^{-1} = \boldsymbol W_{ax} \boldsymbol A \\
\boldsymbol x^{t} = \boldsymbol W_{wx} \boldsymbol w_{t}^{i} + \boldsymbol W_{gx} \boldsymbol G^{t},	
\quad t \in  \{ 0,  \ldots , N^{i}-1 \} \\
\boldsymbol h^{t} = \mathscr R \left( \boldsymbol x^{t} \right),	\quad t \in  \{ -1, 0,  \ldots , N^{i}-1 \} \\
\end{gathered}
\end{equation}
where  \( \boldsymbol W_{ax} \in \mathbb R^{D_{x} \times D_{a}} \) ,  
\( \boldsymbol W_{wx} \in \mathbb R^{D_{x} \times D_{w}} \) , 
\( \boldsymbol W_{gx} \in R^{D_{x} \times D_{g}} \)  are projection matrices of the attribute space, word embedding space and guiding vector space to the LSTM-d input space.  \( D_{x} \), \( D_{w} \), \( D_{a} \), \( D_{g} \) 
denote the dimension of LSTM-d input space, word embedding space, 
attributes vector and guiding vector, respectively. 
The vector  \( \boldsymbol w_{t}^{i} \) 
is the distributed representation of the  \( t \) -th word in caption  \( S^{i} \) . 
We padded each caption with  \( \boldsymbol w_{0}^{i} \)  to the left and 
 \( \boldsymbol w_{N^{i}}^{i} \)  to the right, where  \( \boldsymbol w_{0}^{i} \) 
 and  \( \boldsymbol w_{N^{i}}^{i} \) represent respectively the 
 distributed representations of start-of-sentence 
 token $<$sos$>$ and end-of-sentence token $<$eos$>$. 
 The sequential guiding network (LSTM-g) update its guiding vector  \( \boldsymbol G^{t} \) with the   
 following the procedure: 
 \begin{equation}
\begin{gathered}
\boldsymbol z^{0} = \boldsymbol W_{z} \left[ \boldsymbol h^{-1}; \boldsymbol A \right] \\
\boldsymbol z^{t} = \boldsymbol W_{z} \left[ {\boldsymbol {\tilde h}}^{t-1}; \boldsymbol A \right],
\quad t \in  \{ 1,  \ldots , N^{i}-1 \} \\
\boldsymbol g^{t} = f^{lstm} \left( \boldsymbol z^{t} \right),	\quad t \in  \{ 0,  \ldots , N^{i}-1 \} \\
\end{gathered}
\end{equation}
where \( f^{lstm} \) is the recursive function within a LSTM cell and \( \boldsymbol W_{z} \) is a projection matrix. 

\begin{figure}[ttb]
\centering
\includegraphics[width=0.82\linewidth, height=4.3cm]{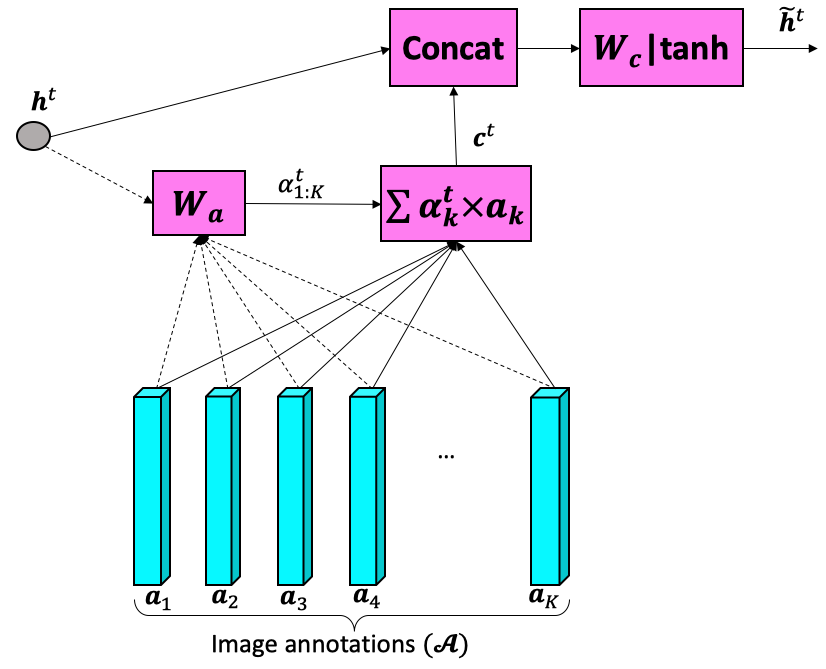}
\caption{Luong's attention applied on image regions.}
\label{fig:atmodel}
\end{figure}

\section{Experimental Studies}
\label{sec:exps}
\subsection{Dataset and Preprocessing}
\label{ssec:ddp}
MS COCO Captions \cite{Lin2014MicrosoftCC, Chen2015MicrosoftCC} is a large scale benchmark dataset for image captioning. We work only with the MS COCO $c5$ dataset which annotates each image with 5 human-produced captions. It contains 82,783 images for training and 40,504 images for validation and 40,775 images for testing. We use the splits publicly available 
\footnote{http://cs.stanford.edu/people/karpathy/deepimagesent/} 
in previous works \cite{Karpathy, Jia2015GuidingLT}. These splits contain 5000 validation images and 5000 testing images taken from the original validation set. The attribute detectors are trained with the same training set and the 1000 most common words in training captions are used to form the attributes. We follow the same data preprocessing in \cite{Karpathy}. That is all captions are transformed to lowercase, non-alphanumeric characters are discarded and all words occurring less than 5 times in the training captions are filtered and replaced by special token $<$unk$>$. The preprocessing results in a vocabulary of 8791 words. 

\subsection{Configurations and Implementation}
\label{ssec:cid}
To obtain image annotations  \( \mathcal{A} \) , we use Inception-V3 \cite{Szegedy2016RethinkingTI} vision model. We take the  \( 8 \times 8 \times 2048 \)  activations map of the last convolutional layer (Mixed\_7c in TensorFlow) as annotations. That is  \( \mathcal{A} \) has dimensions of  \( 64 \times 2048 \) . To avoid overfitting, we did not train the Inception-V3 from scratch but a model pre-trained on ImageNet is used. We did not fine-tune the weights of the vision model though it could give a small performance boost. The dimension of all layers in the decoding LSTM is set to 1024. For the guiding LSTM, the dimension is set to 512. Both LSTM cells are wrapped with dropout to avoid overfitting. We use word vectors with dimension of 512 and initialize all model parameters (except the CNN parameters) randomly with uniform distribution in [-0.1, 0.1]. We built our model based on TensorFlow \cite{Abadi2015TensorFlowLM} and used the publicly available code of Google's \emph{NIC} model as base code. 


\subsection {Model Comparisons}
\label{ssec:ccm}
We validate the effectiveness of our proposed framework by comparing it to several state-of-the-art captioning models based on
CNN+RNN architecture such as NIC \cite{Vinyals2017ShowAT}, 
Soft-Attention \cite{Xu2015ShowAA}, LSTM-A5 \cite{Yao2017BoostingIC}, 
LTG-Soft-Attention \cite{Jiang2018LearningTG} and LTG-Review-Net 
\cite{Jiang2018LearningTG}. We measure the performance of our proposed
model with four popular evaluation metrics: BLEU \cite{bleu}, METEOR \cite{W05-0909}, ROUGE \cite{rouge} 
and CIDEr \cite{Vedantam2015CIDErCI}. To compute  these metrics, we use
the official MS COCO evaluation toolkit\footnote{https://github.com/tylin/coco-caption} that is made publicly available.
Evaluation results across these  metrics are shown in Table \ref{table:results}. Our 
framework is referenced as SGN+Luong-Attention. Note that for all 
models we report only single model performance knowing that ensemble 
of multiple models can always give few extra performance points. As 
shown in the table, our proposed model outperforms all other CNN+RNN 
architectures. It is worth nothing that our model can be seen as an 
extension of all previous methods and is more complex than most of them. However, the use of a sequential guiding network can be given credit by the fact that our proposed method outperforms the models LTG-Soft-Attention and 
LTG-Review-Net which are as complex as ours. In contrast to our model,
in LTG-Soft-Attention and LTG-Review-Net, the learned guiding vector 
was fixed and couldn't be adjusted during words generation. 

To further investigate the usefulness of a sequential guiding network, 
we also compare our SGN+Luong-Attention captioning model with two 
other baseline that we built: Luong-Attention is a CNN+RNN captioning model based solely on Luong attention mechanism and does not have any extra guiding information at all. ATT+Luong-Attention is a version of SGN+Luong-Attention where we discard the guiding LSTM and replace the guiding vector at each time step by the high-level image attributes. It can be seen from the table that our SGN+Luong-Attention captioning model significantly outperforms ATT+Luong-Attention, again giving credit to the sequential guiding network. It is also worth noting that the Luong-Attention captioning model achieves better performance than the Soft-Attention model in \cite{Xu2015ShowAA}, which proves the advantage of using the Luong style of attention mechanism. 

\begin{table}[t!]
\centering
\begin{tabular}{||m{3.8cm}||m{0.5cm}|m{0.5cm}|m{0.5cm}|m{0.7cm}||}
\hline
 Image Captioning Models & B@4 & MTR & RGE & CDr \\ [0.5ex] 
\hline\hline
Google NIC & 20.3 & - & - & - \\
Soft-Attention & 24.3 & 23.9 & - & - \\
LSTM-A5 & 32.5 & 25.1 & 53.8 & 98.6 \\ 
LTG-Soft-Attention & 32.3 & 25.9 & 53.7 & 102.3 \\
LTG-Review-Net & 33.6 & 26.1 & 54.8 & \bf 103.9 \\
\hline\hline
SGN+Luong-Attention (ours) & \bf{34.0} & \bf 26.3 & \bf 55.2 & 103.6 \\
Luong-Attention (ours) & 27.1 & 24.5 & 51.7 & 86.8 \\
ATT+Luong-Attention (ours) & 32.8 & 25.7 & 54.1 & 101.9 \\ [1ex] 
\hline
\end{tabular}
\caption{Single model evaluation results on the 5000 testing images. B@4, MTR, RGE, CDr are respectively short for BLEU@4, METEOR, ROUGE-L and CIDEr.}
\label{table:results}
\end{table}

\section {Conclusion}
\label{sec:con}
In this paper, we have extended the encoder-decoder framework for image captioning by inserting a guiding network. By modeling the guiding network with a Long Short-Term Memory, the guiding vector can be adjusted at each time step based on the current context and high-level image attributes. We have also explored a natural way of applying Luong's attention over image regions and demonstrated its effectiveness. We then combined these two strategies in a single joint model. Experiments were conducted on the MS COCO Captions dataset and showed that the proposed model achieves superior performance over existing models based on the CNN+RNN architecture trained using maximum likelihood estimation. In future work, we will consider to apply this model to cross-modal information retrieval. 

\vfill\pagebreak

\bibliographystyle{IEEEbib}
\bibliography{Template}

\begin{thebibliography}{10}

\bibitem{Vinyals2017ShowAT}
Oriol Vinyals, Alexander Toshev, Samy Bengio, and Dumitru Erhan,
\newblock ``Show and tell: Lessons learned from the 2015 mscoco image
  captioning challenge,''
\newblock {\em IEEE Transactions on Pattern Analysis and Machine Intelligence},
  vol. 39, pp. 652--663, 2017.

\bibitem{Karpathy}
Andrej Karpathy and Li~Fei-Fei,
\newblock ``Deep visual-semantic alignments for generating image
  descriptions,''
\newblock {\em 2015 IEEE Conference on Computer Vision and Pattern Recognition
  (CVPR)}, pp. 3128--3137, 2015.

\bibitem{Xu2015ShowAA}
Kelvin Xu, Jimmy Ba, Ryan Kiros, Kyunghyun Cho, Aaron~C. Courville, Ruslan
  Salakhutdinov, Richard~S. Zemel, and Yoshua Bengio,
\newblock ``Show, attend and tell: Neural image caption generation with visual
  attention,''
\newblock in {\em ICML}, 2015.

\bibitem{Yao2017BoostingIC}
Ting Yao, Yingwei Pan, Yehao Li, Zhaofan Qiu, and Tao Mei,
\newblock ``Boosting image captioning with attributes,''
\newblock {\em 2017 IEEE International Conference on Computer Vision (ICCV)},
  pp. 4904--4912, 2017.

\bibitem{Jia2015GuidingLT}
Xu~Jia, Efstratios Gavves, Basura Fernando, and Tinne Tuytelaars,
\newblock ``Guiding long-short term memory for image caption generation,''
\newblock {\em CoRR}, vol. abs/1509.04942, 2015.

\bibitem{Jiang2018LearningTG}
Wenhao Jiang, Lin Ma, Xinpeng Chen, Hanwang Zhang, and Wei Liu,
\newblock ``Learning to guide decoding for image captioning,''
\newblock in {\em AAAI}, 2018.

\bibitem{Chen2015MindsEA}
Xinlei Chen and C.~Lawrence Zitnick,
\newblock ``Mind's eye: A recurrent visual representation for image caption
  generation,''
\newblock {\em 2015 IEEE Conference on Computer Vision and Pattern Recognition
  (CVPR)}, pp. 2422--2431, 2015.

\bibitem{Kiros2014UnifyingVE}
Ryan Kiros, Ruslan Salakhutdinov, and Richard~S. Zemel,
\newblock ``Unifying visual-semantic embeddings with multimodal neural language
  models,''
\newblock {\em CoRR}, vol. abs/1411.2539, 2014.

\bibitem{capFang}
Hao Fang, Saurabh Gupta, Forrest~N. Iandola, Rupesh~Kumar Srivastava, Li~Deng,
  Piotr Doll{\'a}r, Jianfeng Gao, Xiaodong He, Margaret Mitchell, John~C.
  Platt, C.~Lawrence Zitnick, and Geoffrey Zweig,
\newblock ``From captions to visual concepts and back,''
\newblock in {\em CVPR}, 2015.

\bibitem{Mao2014DeepCW}
Junhua Mao, Wei Xu, Yi~Yang, Jiang Wang, and Alan~L. Yuille,
\newblock ``Deep captioning with multimodal recurrent neural networks
  (m-rnn),''
\newblock {\em CoRR}, vol. abs/1412.6632, 2014.

\bibitem{Donahue2015LongTermRC}
Jeff Donahue, Lisa~Anne Hendricks, Sergio Guadarrama, Marcus Rohrbach,
  Subhashini Venugopalan, Kate Saenko, and Trevor Darrell,
\newblock ``Long-term recurrent convolutional networks for visual recognition
  and description,''
\newblock {\em 2015 IEEE Conference on Computer Vision and Pattern Recognition
  (CVPR)}, pp. 2625--2634, 2015.

\bibitem{Liu2017MATAM}
Chang Liu, Fuchun Sun, Changhu Wang, Feng Wang, and Alan~L. Yuille,
\newblock ``Mat: A multimodal attentive translator for image captioning,''
\newblock in {\em IJCAI}, 2017.

\bibitem{Chen2017SCACNNSA}
Long Chen, Hanwang Zhang, Jun Xiao, Liqiang Nie, Jian Shao, and Tat-Seng Chua,
\newblock ``Sca-cnn: Spatial and channel-wise attention in convolutional
  networks for image captioning,''
\newblock {\em 2017 IEEE Conference on Computer Vision and Pattern Recognition
  (CVPR)}, pp. 6298--6306, 2017.

\bibitem{Bernardi2017AutomaticDG}
Raffaella Bernardi, Ruken Cakici, Desmond Elliott, Aykut Erdem, Erkut Erdem,
  Nazli Ikizler-Cinbis, Frank Keller, Adrian Muscat, and Barbara Plank,
\newblock ``Automatic description generation from images: A survey of models,
  datasets, and evaluation measures,''
\newblock in {\em IJCAI}, 2017.

\bibitem{Cho2014LearningPR}
Kyunghyun Cho, Bart van Merrienboer, Çaglar G{\"u}lçehre, Dzmitry Bahdanau,
  Fethi Bougares, Holger Schwenk, and Yoshua Bengio,
\newblock ``Learning phrase representations using rnn encoder-decoder for
  statistical machine translation,''
\newblock in {\em EMNLP}, 2014.

\bibitem{MT2}
Dzmitry Bahdanau, Kyunghyun Cho, and Yoshua Bengio,
\newblock ``Neural machine translation by jointly learning to align and
  translate,''
\newblock {\em CoRR}, vol. abs/1409.0473, 2014.

\bibitem{Sutskever2014SequenceTS}
Ilya Sutskever, Oriol Vinyals, and Quoc~V. Le,
\newblock ``Sequence to sequence learning with neural networks,''
\newblock in {\em NIPS}, 2014.

\bibitem{Luong2015EffectiveAT}
Thang Luong, Hieu Pham, and Christopher~D. Manning,
\newblock ``Effective approaches to attention-based neural machine
  translation,''
\newblock in {\em EMNLP}, 2015.

\bibitem{NIPS2012_4824}
Alex Krizhevsky, Ilya Sutskever, and Geoffrey~E Hinton,
\newblock ``Imagenet classification with deep convolutional neural networks,''
\newblock in {\em Advances in Neural Information Processing Systems 25}, pp.
  1097--1105. 2012.

\bibitem{Rennie2017SelfCriticalST}
Steven~J. Rennie, Etienne Marcheret, Youssef Mroueh, Jarret Ross, and Vaibhava
  Goel,
\newblock ``Self-critical sequence training for image captioning,''
\newblock {\em 2017 IEEE Conference on Computer Vision and Pattern Recognition
  (CVPR)}, pp. 1179--1195, 2017.

\bibitem{Liu2017ImprovedIC}
Siqi Liu, Zhenhai Zhu, Ning Ye, Sergio Guadarrama, and Kevin Murphy,
\newblock ``Improved image captioning via policy gradient optimization of
  spider,''
\newblock {\em 2017 IEEE International Conference on Computer Vision (ICCV)},
  pp. 873--881, 2017.

\bibitem{Vedantam2015CIDErCI}
Ramakrishna Vedantam, C.~Lawrence Zitnick, and Devi Parikh,
\newblock ``Cider: Consensus-based image description evaluation,''
\newblock {\em 2015 IEEE Conference on Computer Vision and Pattern Recognition
  (CVPR)}, pp. 4566--4575, 2015.

\bibitem{Anderson2016SPICESP}
Peter Anderson, Basura Fernando, Mark Johnson, and Stephen Gould,
\newblock ``Spice: Semantic propositional image caption evaluation,''
\newblock in {\em ECCV}, 2016.

\bibitem{Hochreiter}
Hochreiter Sepp and Schmidhuber Jurgen,
\newblock ``Long short-term memory,''
\newblock {\em Neural Comput.}, pp. 1735--1780, 1997.

\bibitem{Lin2014MicrosoftCC}
Tsung-Yi Lin, Michael Maire, Serge~J. Belongie, Lubomir~D. Bourdev, Ross~B.
  Girshick, James Hays, Pietro Perona, Deva Ramanan, Piotr Doll{\'a}r, and
  C.~Lawrence Zitnick,
\newblock ``Microsoft coco: Common objects in context,''
\newblock in {\em ECCV}, 2014.

\bibitem{Chen2015MicrosoftCC}
Xinlei Chen, Hao Fang, Tsung-Yi Lin, Ramakrishna Vedantam, Saurabh Gupta, Piotr
  Doll{\'a}r, and C.~Lawrence Zitnick,
\newblock ``Microsoft coco captions: Data collection and evaluation server,''
\newblock {\em CoRR}, vol. abs/1504.00325, 2015.

\bibitem{Szegedy2016RethinkingTI}
Christian Szegedy, Vincent Vanhoucke, Sergey Ioffe, Jonathon Shlens, and
  Zbigniew Wojna,
\newblock ``Rethinking the inception architecture for computer vision,''
\newblock {\em 2016 IEEE Conference on Computer Vision and Pattern Recognition
  (CVPR)}, pp. 2818--2826, 2016.

\bibitem{Abadi2015TensorFlowLM}
Mart{\'i}n Abadi, Ashish Agarwal, and et~al.,
\newblock ``Tensorflow: Large-scale machine learning on heterogeneous
  distributed systems,''
\newblock {\em CoRR}, vol. abs/1603.04467, 2015.

\bibitem{bleu}
Kishore Papineni, Salim Roukos, Todd Ward, and Wei-Jing Zhu,
\newblock ``Bleu: a method for automatic evaluation of machine translation,''
\newblock in {\em ACL}, 2002.

\bibitem{W05-0909}
Satanjeev Banerjee and Alon Lavie,
\newblock ``Meteor: An automatic metric for mt evaluation with improved
  correlation with human judgments,''
\newblock in {\em ACL}, 2005, pp. 65--72.

\bibitem{rouge}
Chin-Yew Lin,
\newblock ``Rouge: a package for automatic evaluation of summaries,''
\newblock July 2004.

\end{thebibliography}

\end{document}